%% file: main.tex
\documentclass[letterpaper, 10 pt, conference]{ieeeconf}  

\IEEEoverridecommandlockouts                              

\usepackage{graphics} 
\usepackage{graphicx}
\usepackage{epsfig} 

\usepackage{times} 
\usepackage{amsmath} 
\usepackage{amssymb}  
\usepackage{bm}
\usepackage{amsfonts}
\DeclareMathAlphabet{\mathpzc}{OT1}{pzc}{m}{it}

\usepackage{hyperref}
\usepackage[capitalise,nameinlink]{cleveref}
\usepackage{color}
\usepackage{tabularx}

\usepackage{tikz}
\usepackage{pgfplots}
\usepackage{pgfplotstable}
\usetikzlibrary{spy}
\pgfplotsset{compat=newest}
\ifdefined\figexternalized
    \usetikzlibrary{external}
\fi
\pgfplotsset{
    width=0.38\textwidth,
    scale only axis=true,
    enlargelimits=false,
    every axis plot post/.append style={
        line width=1.0pt,
        font=\footnotesize,
    }, 
    every axis legend/.append style={
        font=\footnotesize,
        rounded corners,
        line width=1.0pt,
    },
    every non boxed x axis/.append style={x axis line style=-},
    every non boxed y axis/.append style={y axis line style=-},
    axis lines=left,
    legend style={fill=none}
}

\usepackage[caption=false,font=footnotesize]{subfig}
\usepackage{float}

\usepackage{xspace}
\newcommand{\eg}{\hbox{\emph{e.g.}}\xspace}
\newcommand{\ie}{\hbox{\emph{i.e.}}\xspace}

\newcommand{\etal}{\hbox{\emph{et al.}}\xspace}

\usepackage[utf8]{inputenc}
\usepackage{pgfplots}

\title{\LARGE \bf
Multi-mode Trajectory Optimization for Impact-aware Manipulation}

\author{Theodoros Stouraitis$^{*, 1, 2}$, Lei Yan$^{*,1,3}$, Jo\~{a}o Moura$^{1}$, Michael Gienger$^{2}$, and Sethu Vijayakumar$^{1}$
\thanks{$^{*}$Denotes equal contribution for both authors.} 
\thanks{E-mail: theodoros.stouraitis@ed.ac.uk, lei.yan@ed.ac.uk}
\thanks{$^{1}$ Authors are with School of Informatics, University of Edinburgh, Edinburgh, U.K.}
\thanks{$^{2}$ Authors are with Honda Research Institute Europe (HRI-EU), Germany.}
\thanks{$^{3}$ Authors are visiting researchers at the Shenzhen Institute for Artificial Intelligence and Robotics for Society (AIRS), The Chinese University of Hong Kong (Shenzhen), China.}
}

\begin{document}

\maketitle
\thispagestyle{empty}
\pagestyle{empty}

\begin{abstract}


The transition from free motion to contact is a challenging problem in robotics, in part due to its hybrid nature. Additionally, 
disregarding the effects of impacts at the motion planning level 
often results in intractable impulsive contact forces. In this paper, we introduce an impact-aware multi-mode trajectory optimization (TO) method that combines hybrid dynamics and hybrid control in a coherent fashion. A key concept is the incorporation of an explicit contact force transmission model in the TO method. This allows the simultaneous optimization of the contact forces, contact timings, continuous motion trajectories and compliance, while satisfying task constraints. We compare our method against standard compliance control and an impact-agnostic TO method in physical simulations. Further, we experimentally validate the proposed method with a robot manipulator on the task of halting a large-momentum object.


\end{abstract}

\section{Introduction}\label{sec:intro}

\input{sections/intro.tex}

\section{Background}\label{sec:background}
\input{sections/background.tex}


\section{Problem formulation}\label{sec:prob_form}
\input{sections/problem_formulation.tex}

\section{Impact-aware Motion Planning}\label{sec:TO_methods}
\subsection{Impact model}\label{subsec:impact_model}
\input{sections/impact_Dyn.tex}

\subsection{Contact force transmission model}\label{subsec:force_transmission_model}

\input{sections/Force_transmission_model.tex}

\subsection{Multi-mode trajectory optimization for hybrid systems}\label{subsec:hybrid_trajectory_optimization}

\input{sections/TO_methods.tex}

\section{Ablation studies and Experiments}\label{sec:results}

\input{sections/results.tex}

\section{Conclusion}\label{sec:conclusion}
\input{sections/conclusion.tex}




\section*{ACKNOWLEDGMENT}
 This work is supported by the Honda Research Institute Europe, the EPSRC CDT in Robotics and Autonomous Systems (EP/L016834/1), EPSRC UK RAI Hub in Future AI and Robotics for Space (FAIR-SPACE: EP/R026092/1) and UK-India Education and Research Initiative (UKIERI DST 2016-17-0152). The authors would like to thank Henrique Ferrolho for his help with the video.


\bibliographystyle{IEEEtran}
\bibliography{reference}  

\end{document}

%% file: sections/intro.tex
Safe and robust robot manipulation under switching dynamics still poses many challenges.
Typically, manipulation tasks require making and breaking contact with objects. This results in challenges in motion planning and control due to, among other factors,  (i) the hybrid nature of the problem~\cite{mason1986mechanics} and  (ii) the uncertainties that arises due to contact dynamics~\cite{bauza2018gp}.

Strikingly humans are not only competent in object manipulation, but prefer to make contact with non-zero velocities 
to achieve the task
faster and smoother~\cite{bennett1994insights}. The two key enablers to realise this are the ability to skillfully switch between free-motion and contact~\cite{flanagan2006control}, and the capacity to  shift between a variety of control mechanisms depending on the stage of the motion and their associated uncertainties~\cite{johansson1992sensory}. 

Recent hybrid Trajectory Optimization (TO) methods in robotics~\cite{toussaint2018differentiable, stouraitis2018dyadic, hogan2018reactive} 
have demonstrated efficient methods for multi-contact manipulation planning. 
Yet, it is not trivial to transfer these behaviours robustly on to the hardware due to the 
challenge of regulating the transitions between free motion and motion in contact, as well as dealing with imprecise timing
of the transition in the reference motions. To address this, a number of hybrid control~\cite{rijnen2015optimal,rijnen2017control} and compliance control~\cite{roveda2015optimal},  
methods have been proposed.
However, given the inherent limitations of the hardware~\cite{haddadin2009requirements}, the impacts that a 
stand-alone controller is capable of dealing with are limited.

\begin{figure}[t]
\centering
\includegraphics[width=0.95\linewidth]{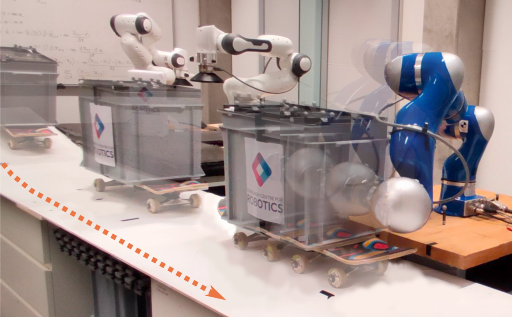}
\caption{Experimental setup where the robot halts an object with a mass of 20 $kg$ travelling at speed of 0.88 $m/s$.}
\label{fig:teaser}
\vspace{-4mm}
\end{figure}

In this work, we try to address this problem at the level of `impact-aware' manipulation planning. 
We ask ourselves, {\it "How could we plan hybrid motions, such that they are easily executable by out-of-the-box controllers?"}, which can be re-framed as a problem of planning such that consistent contact can be maintained during and after impact---even for tasks with contacts at speed, i.e.\ moving objects. 
As a typical example scenario, consider an agent that attempts to stop an object in motion, as shown in \cref{fig:teaser,fig:concept}. In such a case, the agent needs to address the following challenges: 
\begin{itemize}
\item Plan discontinuous motions through contact events and physical impacts, which may result in state triggered velocity jumps described by jump maps \cite{goebel2009hybrid}, i.e., jointly plan continuous motions (flows) and contacts (jumps) to perform a task. 
\item Track discontinuous reference motions, where the actual time of the jumps (impact) may not coincide with the jump time (impact time) of the reference motion. \end{itemize}

A number of motion planning methods have investigated impact related problems. In~\cite{yan2018dual, amanhoud2019dynamical}, non-zero velocities at contact were avoided to exclude impact events.
In~\cite{wang2019impact}, a QP controller 
ensured that unexpected impacts 
will not violate joint limits. They all resulted in either conservative motions or required a priori knowledge of the exact  time of impact.

Catching motions were demonstrated based on learned dynamical systems~\cite{salehian2016dynamical} 
and with TO methods~\cite{lampariello2011trajectory}.
However, the mass of the intercepted objects was negligible and
contact was realized by caging the object. We consider objects with large size and mass, and contacts that can break anytime.

In this paper, we address the two challenges mentioned above with
a coherent contact-invariant TO method that plans `impact-aware' hybrid motions, 
while the control input results from a hybrid controller capable of absorbing impacts. The hybrid controller is based on compliance control that allows to mitigate the peak error due to the mismatch in time between reference and actual impact. Our TO method results in hybrid motions that are inline with the hybrid controller capabilities, while the controller's parameters ({\it e.g.} stiffness) are simultaneously optimized, as in~\cite{rijnen2015optimal}. 
 
The core insight is based on the duality between the impact model used and the capabilities of compliance controllers available in 
the latest collaborative robots.
By modulating the robot's end-effector compliance, we can emulate a number of different types of collisions ranging from elastic to in-elastic, and deduce the optimal force transmission model given the system's limitations, {\it e.g.} workspace limits.

The contributions of our work are:
\begin{itemize}
    \item A parametric programming technique to encode both hybrid dynamics and hybrid control in a single multi-mode   
    trajectory optimization formulation.
    \item  A generic impact model formulation based on a second-order critically damped system to generate smooth contact forces and simultaneously optimize the stiffness. 
    \item A multi-mode  
    trajectory optimization framework that can deal with both multi-contact motion planning and contact force generation for impact-aware manipulation.
\end{itemize}

This paper is organized as follows: 
\cref{sec:background} provides the background on systems with hybrid dynamics and hybrid control and reviews related work on hybrid motion planning. The details of the proposed contact force transmission model and TO method are given in \cref{sec:TO_methods}.
\cref{sec:results} presents the evaluation of the method and the experimental results. Finally, \cref{sec:conclusion} discusses current limitations and future directions.

%% file: sections/background.tex
\subsection{Hybrid dynamic systems}
\label{subsec:hybrid-dym}
As described in \cref{sec:intro}, 
motion planning concepts for manipulation are often based on trajectories that guide an object to its desired state.
In this work, we consider a class of systems where the trajectories include discontinuous transitions between different contact states. Similar to \cite{toussaint2018differentiable, marcucci2017approximate}, we describe systems with hybrid dynamics as 
\begin{equation}
    \label{eq:MPMPC_system_model}
    \dot{\bm{x}}(t) = \bm{f}_k \left(\bm{x}(t), \bm{u}(t), \bm{v}(t)\right), ~\text{if}~ (\bm{x}(t), \bm{u}(t)) \in \mathcal{D}_k,
\end{equation}
where $\bm{x} (t)\in \mathbb{R}^n$ is the state of the system, $\bm{u} (t)\in \mathbb{R}^m$ is the control actions of the plant, $\bm{v} (t)\in \mathbb{R}^\nu$ is the control input applied on the environment, $n,m,\nu \in \mathbb{R}$ define the dimensions of each quantity and $k \in \{0,1\}$ indexes to the different sets $\mathcal{D}_k$. Each $\mathcal{D}_k \subset \mathbb{R}^{n \times m}$ defines the domain (relative to $\bm{x}(t)$ and $\bm{u}(t)$) of a contact state, \ie free-motion or in-contact. Note that \eqref{eq:MPMPC_system_model} defines both the plant's and environment's dynamics. 

For dynamic robot manipulation with contact changes, a number of TO methods~\cite{stouraitis2018dyadic, onol2019contact, sleiman2019contact} have been proposed. The underlying formulations have been borrowed from the locomotion domain~\cite{ posa2014direct, winkler2018gait} 
and can be separated into two classes.
The {\it contact-implicit}~\cite{posa2014direct} and the {\it multi-phase}~\cite{winkler2018gait} or {\it multi-mode}~\cite{toussaint2018differentiable} approaches.  
The former requires special attention 
to the relaxation of the problem to avoid spurious local minima~\cite{nurkanoviclimits}. The latter enables us to obtain a smooth NLP~\cite{nurkanoviclimits} given a mode sequence, which can be obtained from an outer-level process~\cite{toussaint2018differentiable}. In~\cite{toussaint2018differentiable},  each mode is associated with the "contact activity" (physical interaction between objects), is specified 
via path constraints and furthermore, the modes used can be {\it contact}, {\it kinematic} and {\it stable}.
Here, we adopt the latter paradigm as the sequence of modes is fixed and it admits a general notion of modes~\cite{toussaint2018differentiable} not limited to on / off contact.

TO methods for contact planning transcribe the state and the control input of the system, which entails that forces are optimization variables too. Although \cite{patel2019contact} indicated that simply planning contact forces based only on the contact state does not suffice towards making stable contact and pointed out the importance of an accurate force model during contact transitions, most previous works neglected this aspect of the problem. Typically, a number of  assumptions were made to transfer the motion plans to the robots. In~\cite{sleiman2019contact},  purely inelastic collision was assumed to impose no-rebound condition, while in~\cite{onol2019contact}, a variable smooth contact model was used that allows virtual forces from distance. Thus, a natural
design question arises regarding the choice of the contact force transmission model. Such a model can be used to constrain the control inputs and could also be conditioned on the current mode of the system.

\begin{figure}[t]
    \centering
    \def\svgwidth{0.95\linewidth}
    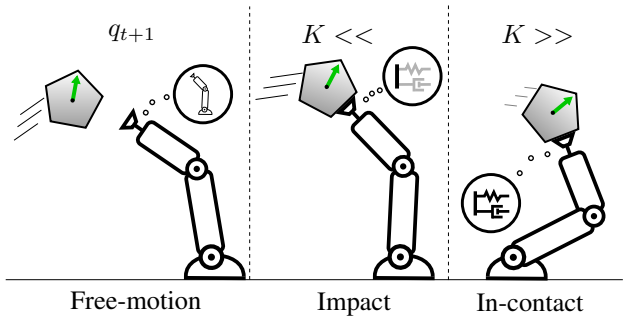
    \caption{Pictorial description of the multi-mode TO for the task of halting a moving object.}\label{fig:concept}
    \vspace{-2mm}
\end{figure}

\subsection{Hybrid control systems}
\label{subsec:hybrid-control}
Hybrid control is useful when dealing with systems that require multiple separate controllers, \eg Ott \etal \cite{ott2010unified} proposed a hybrid force controller for switching between impedance and admittance controllers. In this work, different control modes are used for the deformation and the restitution phases, shown in \cref{fig:impact_model}.
Similar to \cite{hogan2018reactive, utkin2013sliding}, we describe systems with hybrid control as 
\begin{equation}
    \label{eq:MPMPC_control_model}
    \bm{v}(t) = \bm{h}_l \left(\bm{u}(t)\right), 
    ~\text{if}~  (\bm{u}(t)) \in \mathcal{T}_l,
\end{equation}
where $l \in \mathbb{Z}$ indexes to a selected set $\mathcal{T}_l \subset \mathbb{R}^m$.
Each $\mathcal{T}_l$ corresponds to a controller type, \eg impedance, admittance, direct force-control.
\eqref{eq:MPMPC_control_model} specifies the transformation from the plant's control actions to the environment's control inputs.

%% file: figures/concept/catchConcept_small.pdf_tex
\begingroup%
  \makeatletter%
  \providecommand\color[2][]{%
    \errmessage{(Inkscape) Color is used for the text in Inkscape, but the package 'color.sty' is not loaded}%
    \renewcommand\color[2][]{}%
  }%
  \providecommand\transparent[1]{%
    \errmessage{(Inkscape) Transparency is used (non-zero) for the text in Inkscape, but the package 'transparent.sty' is not loaded}%
    \renewcommand\transparent[1]{}%
  }%
  \providecommand\rotatebox[2]{#2}%
  \newcommand*\fsize{\dimexpr\f@size pt\relax}%
  \newcommand*\lineheight[1]{\fontsize{\fsize}{#1\fsize}\selectfont}%
  \ifx\svgwidth\undefined%
    \setlength{\unitlength}{607.38842268bp}%
    \ifx\svgscale\undefined%
      \relax%
    \else%
      \setlength{\unitlength}{\unitlength * \real{\svgscale}}%
    \fi%
  \else%
    \setlength{\unitlength}{\svgwidth}%
  \fi%
  \global\let\svgwidth\undefined%
  \global\let\svgscale\undefined%
  \makeatother%
  \begin{picture}(1,0.50459454)%
    \lineheight{1}%
    \setlength\tabcolsep{0pt}%
    \put(0,0){\includegraphics[width=\unitlength,page=1]{./figures/concept/catchConcept_small.pdf}}%
    \put(0.10508391,0.01046745){\color[rgb]{0,0,0}\makebox(0,0)[lt]{\lineheight{1.25}\smash{\begin{tabular}[t]{l}Free-motion\end{tabular}}}}%
    \put(0.50494814,0.01039866){\color[rgb]{0,0,0}\makebox(0,0)[lt]{\lineheight{1.25}\smash{\begin{tabular}[t]{l}Impact\end{tabular}}}}%
    \put(0.76283114,0.00802289){\color[rgb]{0,0,0}\makebox(0,0)[lt]{\lineheight{1.25}\smash{\begin{tabular}[t]{l}In-contact\end{tabular}}}}%
    \put(0,0){\includegraphics[width=\unitlength,page=2]{./figures/concept/catchConcept_small.pdf}}%
    \put(0.16596728,0.45327285){\color[rgb]{0,0,0}\makebox(0,0)[lt]{\lineheight{1.25}\smash{\begin{tabular}[t]{l}$q_{t+1}$\end{tabular}}}}%
    \put(0,0){\includegraphics[width=\unitlength,page=3]{./figures/concept/catchConcept_small.pdf}}%
    \put(0.80324337,0.44308165){\color[rgb]{0,0,0}\makebox(0,0)[lt]{\lineheight{1.25}\smash{\begin{tabular}[t]{l}$K>>$\end{tabular}}}}%
    \put(0.47963847,0.44308165){\color[rgb]{0,0,0}\makebox(0,0)[lt]{\lineheight{1.25}\smash{\begin{tabular}[t]{l}$K<<$\end{tabular}}}}%
  \end{picture}%
\endgroup%

%% file: sections/problem_formulation.tex
One can observe from \eqref{eq:MPMPC_system_model} and \eqref{eq:MPMPC_control_model}, that the investigated system has a variety of different contact states and controllers that can alter the system's behaviour along the time axis. We refer to a single combination of a contact state and a controller as a mode of the system. The proposed notion for contact-control modes is similar to the notion of physical interaction modes introduced in~\cite{toussaint2018differentiable} (see \cref{subsec:hybrid-dym}). Here, we only consider a limited number of contact states as physical interaction modes, but we extend the notion of mode by considering a variety of different controllers.

The sequential arrangement of these modes $z_j = \{ (k_j,l_j)\}$ 
defines the outline of the trajectory, while for each different sequence of contact-control modes $\bm{z}: \{ z_0, z_1, ... z_J \}$ there is a different optimal solution of state $ ^*\bm{x} (t)$ and control $ ^*\bm{u}(t)$ trajectories. $J \in \mathbb{Z}^+$ describes the total number of modes of the trajectory.

Given a mode sequence, the multi-mode trajectories described by \eqref{eq:MPMPC_system_model} and \eqref{eq:MPMPC_control_model} can be explicitly expressed as a function of the initial state and the plant's action sequence. 
Inspired by \cite{toussaint2018differentiable, hogan2018reactive, marcucci2017approximate}, we think of impact-aware manipulation planning
as a special form of Parametric Programming (PP)~\cite{multi-parametricProg}, where the sequence of modes $\bm{z}$ is encoded in the problem as
\begin{IEEEeqnarray}{CCCCC}
    \IEEEyesnumber\label{eq:csDSTO} 
    \IEEEyessubnumber \label{eq:csDSTO_cost}
    \min_{\bm{x}(t), \bm{u}(t), \bm{v}(t)}  &  \bm{c}\left(\bm{x}(t), \bm{u}(t), \bm{v}(t), \bm{z}\right)~ \\
    ~\text{s.t.}~  
    \IEEEyessubnumber \label{eq:csDSTO_system_model} &
    \dot{\bm{x}}(t) = \bm{f} \left(\bm{x}(t), \bm{u}(t), \bm{v}(t), \bm{z}\right),~\hfill\\
    \IEEEyessubnumber \label{eq:csDSTO_control_model} & 
    \dot{\bm{v}}(t) = \tilde{\bm{h}} \left(\bm{u}(t), \bm{z}\right), \hfill\\
    \IEEEyessubnumber \label{eq:csDSTO_constraints} & 
    \bm{g}(\bm{x}(t), \bm{u}(t), \bm{v}(t), \bm{z}) \leq 0. \hfill
\end{IEEEeqnarray}
\eqref{eq:csDSTO_cost} - \eqref{eq:csDSTO_constraints} are piecewise functions from which the appropriate piece (interval) can be selected based on $\bm{z}$. \eqref{eq:csDSTO_cost} defines the objective function, and~$\bm{g}(\cdot)$ in \eqref{eq:csDSTO_constraints} represents both the equality and the inequality constraints of the system. It is worth pointing out that Optimal Control (OC)  problems with hybrid dynamics are usually written as in \eqref{eq:csDSTO}, excluding \eqref{eq:csDSTO_control_model}, while OC problems with hybrid control are usually written as in \eqref{eq:csDSTO}, excluding \eqref{eq:csDSTO_system_model}. The formulation above defines an OC problem where both dynamics and control are hybrid. Further, we enforce \eqref{eq:MPMPC_control_model} as a dynamical system through \eqref{eq:csDSTO_control_model}. The details on this decision are given in the next section. 

Next, we consider one instantiation of such a problem---characterised as halting a moving object. For this task, the robot has to be initially soft to absorb the impact and then, stiff to accurately manipulate the object.

%% file: sections/impact_Dyn.tex
In scenarios where two objects collide (contact transition) with non-zero relative velocity, an impulsive force results. The velocity discontinuity between pre-impact and post-impact is described with the following relationship
\begin{equation}
    M \left(\bm{\mathrm{v}}^+ - \bm{\mathrm{v}}^- \right) = \bm \Lambda \delta t,
\end{equation}
where $M$ is the mass of the system, $\bm{\mathrm{v}}^-$ and $\bm{\mathrm{v}}^+$ are the pre-impact and post-impact relative velocities, respectively. $\bm \Lambda$ is the impact force and $\delta t \simeq 0$ is the impact's duration.

For a moving object that experiences an impact, 
the dissipated energy---due to the impulsive force---during the collision is
\begin{equation}
    \label{eq:impactEnergy}
    E_{\Lambda} = \frac{1}{2}M{\mathrm{v}^-}^2 - \frac{1}{2}M{\mathrm{v}^+}^2.
\end{equation} 

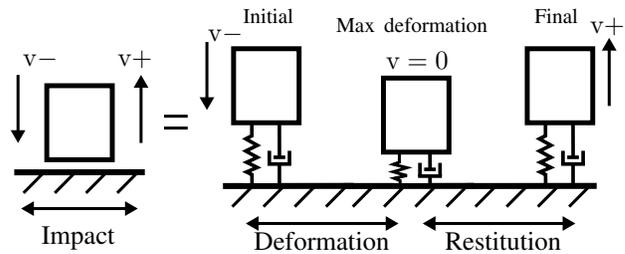
\begin{figure}[t]
    \centering
    \def\svgwidth{0.95\linewidth}
    \input{./figures/background/impact_model.tex}
    \caption{Correspondence between Newton's restitution model and the mass-spring-damper system.}
    \label{fig:impact_model}
    \vspace{-3mm}
\end{figure}

In this paper, we adopt the mass-spring-damper system to model real-world collisions~\cite{nagurka2004mass}. The equation of motion for a mass-spring-damper system shown in \cref{fig:impact_model} is written as 
\begin{equation}\label{eq:mass_spring_damper}
    M \ddot{x} +B \dot{x} + K x = -Mg,
\end{equation}
where $K, B, M$ are the stiffness, damping and mass respectively;
$g$ is gravity and $x$ is the state of the system. 
The energy dissipation of such a system is caused by the damper and can be calculated as follows
\begin{equation}
    \label{eq:mass_spring_damperEnergy}
    E_p = \int_0^{\delta t} B\dot{x}dx = \int_0^{\delta t} B \dot{x}^2 dt.
\end{equation}
Thus, by equating \eqref{eq:impactEnergy} with \eqref{eq:mass_spring_damperEnergy}, we can model the energy loss during impact with a spring-damper system~\cite{nagurka2004mass}, where the dissipated energy during deformation and restitution stages (see \cref{fig:impact_model}) is related to both stiffness and damping.

In addition, based on~\cite{nagurka2004mass, zhu1999theoretical}, the characteristics of the physical system such as duration of impact and restitution coefficient\footnote{The restitution coefficient value $\epsilon_r=1$ represents a perfectly elastic collision,~$0\leq\epsilon_r\leq 1$ represents a real-world inelastic collision. ~$\epsilon_r=0$ represents a perfectly inelastic collision.} can be related to the mass, damping and stiffness parameters of the mechanical system.
We utilize this observation to accurately emulate the physical interaction through the impedance controller of the manipulator. 
As shown in Fig.~\ref{fig:impact_model}, the negative contact is defined as the deformation stage during which the contact force for making a stable contact is generated. The positive contact is defined as the restitution stage which  generates the contact force for the manipulation tasks, such as pushing an object away. We encode these two stages of the contact as
\begin{equation}\label{eqn:contact_phases}
    l = \left\{ \begin{array}{cc}
        -1, & \bm{\mathrm{v}}^- \rightarrow 0\\
        1, & 0 \rightarrow \bm{\mathrm{v}}^+.
    \end{array}\right.
\end{equation}

In terms of impact-aware manipulation, the stiffness should be minimized during negative contact while it should be maximized during positive contact, to achieve accurate manipulation. These contact stages are encoded in \eqref{eq:csDSTO} in the form of controllers according to  \eqref{eq:MPMPC_control_model}. 
In this way, the controller parameters (stiffness) are optimized to conform with the different stages of the contact. 

%% file: figures/background/impact_model.tex
\begingroup%
  \makeatletter%
  \providecommand\color[2][]{%
    \errmessage{(Inkscape) Color is used for the text in Inkscape, but the package 'color.sty' is not loaded}%
    \renewcommand\color[2][]{}%
  }%
  \providecommand\transparent[1]{%
    \errmessage{(Inkscape) Transparency is used (non-zero) for the text in Inkscape, but the package 'transparent.sty' is not loaded}%
    \renewcommand\transparent[1]{}%
  }%
  \providecommand\rotatebox[2]{#2}%
  \newcommand*\fsize{\dimexpr\f@size pt\relax}%
  \newcommand*\lineheight[1]{\fontsize{\fsize}{#1\fsize}\selectfont}%
  \ifx\svgwidth\undefined%
    \setlength{\unitlength}{462.75bp}%
    \ifx\svgscale\undefined%
      \relax%
    \else%
      \setlength{\unitlength}{\unitlength * \real{\svgscale}}%
    \fi%
  \else%
    \setlength{\unitlength}{\svgwidth}%
  \fi%
  \global\let\svgwidth\undefined%
  \global\let\svgscale\undefined%
  \makeatother%
  \begin{picture}(1,0.41329011)%
    \lineheight{1}%
    \setlength\tabcolsep{0pt}%
    \put(0,0){\includegraphics[width=\unitlength,page=1]{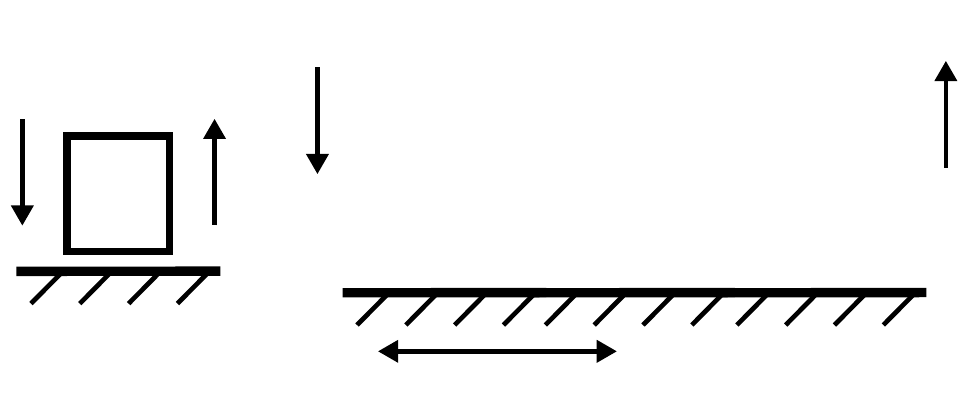}}%
    \put(0.06107727,0.01665933){\color[rgb]{0,0,0}\makebox(0,0)[lt]{\lineheight{1.25}\smash{\begin{tabular}[t]{l}Impact\end{tabular}}}}%
    \put(0,0){\includegraphics[width=\unitlength,page=2]{./figures/background/impactModel.pdf}}%
    \put(0.3878773,0.37330579){\color[rgb]{0,0,0}\makebox(0,0)[lt]{\lineheight{1.25}\smash{\begin{tabular}[t]{l}{\footnotesize Initial}\end{tabular}}}}%
    \put(0.40586694,0.00641533){\color[rgb]{0,0,0}\makebox(0,0)[lt]{\lineheight{1.25}\smash{\begin{tabular}[t]{l}Deformation\end{tabular}}}}%
    \put(0,0){\includegraphics[width=\unitlength,page=3]{./figures/background/impactModel.pdf}}%
    \put(0.53879829,0.35942269){\color[rgb]{0,0,0}\makebox(0,0)[lt]{\lineheight{1.25}\smash{\begin{tabular}[t]{l}{\footnotesize Max} {\footnotesize deformation}\end{tabular}}}}%
    \put(0,0){\includegraphics[width=\unitlength,page=4]{./figures/background/impactModel.pdf}}%
    \put(0.86025899,0.37330579){\color[rgb]{0,0,0}\makebox(0,0)[lt]{\lineheight{1.25}\smash{\begin{tabular}[t]{l}{\footnotesize Final}\end{tabular}}}}%
    \put(0,0){\includegraphics[width=\unitlength,page=5]{./figures/background/impactModel.pdf}}%
    \put(0.71543317,0.00641533){\color[rgb]{0,0,0}\makebox(0,0)[lt]{\lineheight{1.25}\smash{\begin{tabular}[t]{l}Restitution\end{tabular}}}}%
    \put(0.03084818,0.3048104){\color[rgb]{0,0,0}\makebox(0,0)[lt]{\lineheight{1.25}\smash{\begin{tabular}[t]{l}$\mathrm{v}-$\end{tabular}}}}%
    \put(0.18484835,0.3048104){\color[rgb]{0,0,0}\makebox(0,0)[lt]{\lineheight{1.25}\smash{\begin{tabular}[t]{l}$\mathrm{v}+$\end{tabular}}}}%
    \put(0.33105439,0.34457444){\color[rgb]{0,0,0}\makebox(0,0)[lt]{\lineheight{1.25}\smash{\begin{tabular}[t]{l}$\mathrm{v}-$\end{tabular}}}}%
    \put(0.95012152,0.35702789){\color[rgb]{0,0,0}\makebox(0,0)[lt]{\lineheight{1.25}\smash{\begin{tabular}[t]{l}$\mathrm{v}+$\end{tabular}}}}%
    \put(0.61967327,0.29990187){\color[rgb]{0,0,0}\makebox(0,0)[lt]{\lineheight{1.25}\smash{\begin{tabular}[t]{l}$\mathrm{v} = 0$\end{tabular}}}}%
  \end{picture}%
\endgroup%

%% file: sections/Force_transmission_model.tex
For a smooth transition from free-motion to contact, the contact duration and contact force profile should obey
the impact model shown in~\cref{fig:impact_model}. 
As such~\eqref{eq:mass_spring_damper} becomes 
\begin{equation}\label{eqn:impact_force_model}
    M \Delta \ddot{c} + B \Delta \dot{c} + K \Delta c =  \mathrm{f}_d,
\end{equation}
where $\mathrm{f}_d$ is the desired contact force, $\Delta c$, $\Delta \dot{c}$, and $\Delta \ddot{c}$ are the deformed position, velocity and acceleration.

In order to plan smooth contact force without oscillations, we model the force transmission as a second-order critically damped dynamical system (cd-DS). 
A cd-DS~\cite{braun2013robots} was first used 
to guarantee that the motor position is tractable, and it was further used to provide constraint consistent output for any admissible input.
In this paper, we formulate a cd-DS for contact force transmission as 
\begin{equation}\label{eqn:critically_damped_differential_equation}
\ddot {\mathrm{f}}(t) + 2 \alpha \dot {\mathrm{f}}(t) + \alpha^2 {\mathrm{f}(t)} = \alpha^2 {\mathrm{f}}_d,
\end{equation}
where the contact force $\mathrm{f}(t)$ satisfies $\mathrm{f}(t) \in [0, \mathrm{f}_d]$, while ${\dot{\mathrm{f}}(t)}$ and ${\ddot{\mathrm{f}}(t)}$ are its first and second derivatives. For any $ {\alpha} >0$, the contact force $\mathrm{f}(t)$ is critically damped.

Additionally, we enforce the second-order contact force transmission model \eqref{eqn:critically_damped_differential_equation} to have the same characteristics as the impact model \eqref{eqn:impact_force_model}, i.e. the same natural frequency $\omega_n$ and damping ratio $\mathrm{\zeta}$.
Give that the $K {\Delta \bm c} = \mathrm{f}, {\Delta \bm \dot{c}} \approx 0$ and ${\Delta \bm \ddot{c}} \approx 0$ at the maximum deformation point of \eqref{eqn:impact_force_model} and the
Laplace transform of \eqref{eqn:impact_force_model} and \eqref{eqn:critically_damped_differential_equation}, 
\begin{equation}\label{eqn:admittance_transfer_function}
{\Delta c} = \frac{1}{ M s^2+  B s+  K} \mathrm{f}_d  = \frac{\frac{K}{M}}{ s^2+  \frac{B}{M} s+ \frac{K}{M}} \frac{\mathrm{f}_d}{K},
\end{equation}
\begin{equation}\label{eqn:cdDS_transfer_function}
{\mathrm{f}}(t) = \frac{ {\alpha}^2}{s^2+2 {\alpha} s + {\alpha}^2}   {\mathrm{f}}_d,
\end{equation}
we can obtain the following relationship between parameter $\alpha$ and the 
parameters of the impact model~\eqref{eqn:impact_force_model}:\\

\begin{minipage}{.45\linewidth}
\begin{equation}\label{eqn:mk2alpha}
    {\alpha} = \sqrt{\frac{ K}{M}},
    \end{equation}
\end{minipage}%
\begin{minipage}{.5\linewidth}
\begin{equation}\label{eqn:mk2b}
    B = 2\sqrt{M K}.
    \end{equation}
\end{minipage}\\

Thus, given the mass of the object and the stiffness parameter we can obtain the cd-DS parameter $\alpha$. Further, for the differential equation~\eqref{eqn:critically_damped_differential_equation} with output $\mathrm{f}(t)$, input $\mathrm{f}_d$ and damping factor $\mathrm{\zeta}=1$, we can obtain the  relationship between $\alpha$ and settling time $t_s$ (within 5\%, $t_s \approx \frac{3.0}{\omega_n \mathrm{\zeta}} = \frac{3.0}{\alpha}$).
This reveals that $\alpha$ is coupled to the contact duration.
Also, the feasible contact duration is related to the velocity of object and the workspace of the robot; thus, $\alpha$
should be limited according to the attributes of the physical system.
In the multi-mode TO (see next section), both $\alpha$---\ie stiffness---and contact duration are optimized to satisfy the workspace limits of the robot.

Simultaneously, the mass-spring-damper system which establishes the relationship between force and position is also adopted in the impedance control to regulate the operational force of the manipulator. To achieve
fast tracking performance without oscillation with such a controller, the mass, damping and stiffness parameters of ~\eqref{eqn:impact_force_model} need to form a second-order critically damped system. 
Thus, based on the optimized parameter $\alpha$, the optimal inertia, damping and stiffness of the impedance controller are obtained from~\eqref{eqn:mk2alpha} and~\eqref{eqn:mk2b} which satisfy the critically damped constraint. 
This enables impact-aware contact transitions and fast contact force tracking performance without oscillations.

%% file: sections/TO_methods.tex
To solve the continuous optimization problem in \eqref{eq:csDSTO}, we discretize the trajectory according to direct transcription~\cite{rawlings2017model}.
The transcription of our hybrid parametric optimization problem is an extension of the phase-based parameterization used in our previous work~\cite{stouraitis2018dyadic} and is similar in spirit to~\cite{toussaint2018differentiable}. For each $i_{th}$ knot\footnote{Knots are the discretization points of the transcribed continuous problem.}, the decision variables are (i) the pose of the object $\mathbf{y}_i$, (ii) the velocity of the object $\mathbf{\dot{y}}_i \in \mathbb{R}^\nu$,  (iii) action timings $\Delta \mathbf{T}_i$, (iv) the end-effector's position $\mathbf{c}_i$, (v) the contact force $\mathbf{f}_i$
and the cd-DS parameter $ \bm{\alpha}$. We group these quantities into three vectors 
\begin{IEEEeqnarray}{C}
    \IEEEyesnumber\label{eq:OptVariables}
    \bm{x}_i = \begin{bmatrix}\mathbf{y}_i & \mathbf{\dot{y}}_i & \mathbf{c}_i & \mathbf{\dot{c}}_i & \mathbf{\ddot{c}}_i \end{bmatrix}^T_, \\
    \bm{u}_i = \begin{bmatrix} \bm{\alpha}_i & \Delta \mathbf{T}_i \end{bmatrix}^T_, \\
    \bm{v}_i = \begin{bmatrix} \mathbf{f}_i & \mathbf{\dot{f}}_i & \mathbf{\ddot{f}}_i  \end{bmatrix}^T_,
\end{IEEEeqnarray}
where $\forall i \in  \mathbb{N}$, the trajectories of $\bm{x}_i$, $\bm{u}_i$ and $\bm{v}_i$ describes a multi-mode motion.  
In addition to the decision variables, 
the transcription of the continuous problem can be customized through the mode sequence $\bm{z}$.
This results in a TO problem that is separated into modes with different constraints.

\subsubsection*{\textbf{Mode-free constraints}}

Here we introduce all the constraints that are applied independently to the modes of the  trajectory, i.e., constraints that are free of parameter set $\bm{z}$. We note that $\psi_c \in  \mathbb{R}^{2\nu}$ defines the reachable area of the agent's end-effectors, referred to as workspace.

\begin{itemize}
   
    \item Initial state of the object: $  \mathbf{y}_{0} = \mathbf{y}_0^* \text{ and } \mathbf{\dot{y}}_{0} = \mathbf{\dot{y}}_0^* $.
  
    \item Desired final state of the object: $ \mathbf{y}_{N}  = \mathbf{y}_N^*$  and / or  $ \mathbf{\dot{y}}_{N} = \mathbf{\dot{y}}_N^* $.

    \item Kinematic limits of the end-effector: $\mathbf{c}_i \in \psi_c \label{eq:boxconstraint} $, approximated with box bounds.

    \item Lower and upper bound on time between each knot: $ \Delta T^{l} \leq  \Delta T_i \leq \Delta T^{u}, ~ \forall i \in \{0,...,N\} $.  
\end{itemize}

\subsubsection*{\textbf{Mode-conditioned
constraints}}\label{subsubsec:phaseparameterization}
The set of contact-control modes allows to specify the subset of constraints needed to be satisfied at each discretization point (knot). Here, we organize the description of the constraints into time-dependent and time-independent constraints, respectively.

\paragraph{Time-dependent}
\eqref{eq:csDSTO_system_model} describes the dynamics of the system, which includes both the dynamics of the object and the motion of the end-effector. We note that the object's dynamics $f_o(\cdot) \in  \mathbb{R}^{2\nu}$ and the end-effectors' motion $f_e(\cdot) \in  \mathbb{R}^{\nu}$ are integrated using trapezoidal quadrature. Additionally, \eqref{eq:csDSTO_system_model} depends on the current mode and therefore, will be written in a piecewise form. 

\paragraph*{\underline{Object's dynamics}}
As discussed in \cref{sec:background}, when $k=0$, the object is in free motion while when $k=1$, the object is in contact. According to the above, the dynamics of the object are described by
\begin{align}
\label{eq:objDyn}
\begin{bmatrix} \mathbf{y}_{i+1} \\ \mathbf{\dot{y}}_{i+1}\end{bmatrix} = \left\{ \begin{array}{ll} 
        \bm{f}_o(\mathbf{y}_i, \mathbf{\dot{y}}_i, \mathbf{c}_i, \Delta \mathbf{T}_i), & \text{if $k_i=0$},\\
        \bm{f}_o(\mathbf{y}_i, \mathbf{\dot{y}}_i, \mathbf{c}_i, \Delta \mathbf{T}_i, \mathbf{f}_i), & \text{if $k_i=1$},
        \end{array} \right.
\end{align}
while \eqref{eq:objDyn} for $k=1$ in more detail is
\begin{equation}
    \label{eq:FullDyn}
    \begin{bmatrix} M \bm I & 0 \\ 0 & \bm J \end{bmatrix} \mathbf{\ddot{y}}_i + \begin{bmatrix} M \bm g \\ \mathbf{\dot{y}}^\omega_i \times (\bm J \mathbf{\dot{y}}^\omega_i) \end{bmatrix} = \begin{bmatrix} \bm I \\ \hat{\mathbf{c}}_i \end{bmatrix} \mathbf{f}_i,
\end{equation} 
where $ M \in \mathbb{R}$  and $\bm J \in \mathbb{R}_{\ge 0}^{\nu \times \nu}$ are the mass and inertia of the object, \(\bm I\) is the identity matrix, $\bm g$ is the acceleration due to gravity, $\mathbf{\dot{y}}^\omega_i$ is the object's angular velocity, and we refer to the cross product matrix formed by the input vector with \(\hat{(\cdot)}\). With \eqref{eq:FullDyn}, the hybrid nature of the system's dynamics is evident (see \cref{subsec:hybrid-dym}). If $k=1$ the RHS of \eqref{eq:FullDyn} remains while if $k=0$, the RHS of \eqref{eq:FullDyn} disappears.

\paragraph*{\underline{End-effector's motion}}
When planning motions with impacts, particular care needs to be taken while enforcing the integration constraints of the motion \cite{goebel2009hybrid}. The motion of the end-effector is described with the following function 
\begin{equation}
\label{eq:eeMotion}
\begin{bmatrix} \mathbf{c}_{i+1} \\ \mathbf{\dot{c}}_{i+1}\end{bmatrix} = \left\{ \begin{array}{l} 
                f_e(\mathbf{c}_i, \mathbf{\dot{c}}_i, \mathbf{\ddot{c}}_i, \Delta \mathbf{T}_i), \text{ if $k_i = 0$ or $1 $}~\\
                f_e(\mathbf{c}_i, \mathbf{\dot{c}}_i, \Delta \mathbf{T}_i), \text{ if $k_i=0,$ $k_{i+1}=1$}~\\
                f_e(\mathbf{c}_i, \mathbf{\dot{c}}_i, \Delta \mathbf{T}_i), \text{ if $k_i=1,$ $k_{i+1}=0$}~ 
                \end{array} \right.
\end{equation}
where time integration from accelerations to velocities needs to be skipped at specific mode transitions, similar to \cite{rijnen2017control}. These transitions are the making and the breaking of contact, which are liable to impact and are noted by $k_{i} = 0, k_{i+1} = 1$ and $k_{i} = 1, k_{i+1} = 0$, respectively. By omitting $\mathbf{\ddot{c}}$ at these transitions, velocity jumps are possible. Hence, the space of solutions of the mathematical program includes state jumps, typically described with jump maps \cite{goebel2009hybrid}. Note that we need to omit the time integration of the plant, but not the one of the object.  
The forces in \eqref{eq:objDyn} are unbounded (can trigger velocity jumps), while the accelerations of the plant need to be bounded---which constrains velocity changes (curb velocity jumps)---according to the capabilities of the plant.

\paragraph*{\underline{Contact force transmission}}
Based on the impact model and the contact force transmission model, 
the forces of the TO are generated according to the following differential equation 
\begin{equation}
    \label{eq:ControlDyn}
    \begin{bmatrix} \mathbf{\dot{f_{i}}}  \\  \mathbf{\ddot{f_{i}}} \end{bmatrix} =
    \begin{bmatrix} 0 & 1 \\ -{\bm \alpha_l}^2 & -2{\bm \alpha_l} \end{bmatrix} \begin{bmatrix} \mathbf{f_{i}} \\ \mathbf{\dot{f_{i}}} \end{bmatrix} + \begin{bmatrix} 0 \\ {\bm \alpha}_l^2 \end{bmatrix} \mathbf{f}_d.
\end{equation}
According to \eqref{eq:ControlDyn}, the contact force transmission model is parameterized by only one parameter $\bm \alpha_l$.
Hybrid control (see \cref{subsec:hybrid-control}) is realized with two controllers (i) $l=-1$ and (ii) $l=1$, one for each contact stage defined in \eqref{eqn:contact_phases}. For each controller, 
$\bm \alpha_l$ is optimized to modulate the contact time and contact force profile.
Furthermore, based on the relationship between $\alpha$ and $K$ in \eqref{eqn:mk2alpha}, the stiffness characteristics are also optimized in a coherent way through $\bm \alpha_l$---without separating the contact scheduling from stiffness modulation into two levels as in~\cite{nakanishi2013spatio}.

\paragraph{Time-independent}

These constraints are grouped with respect to the modes of the trajectory.

\textit{\underline{Free-motion mode ($k=0$)}}:\\
\indent \indent $\bullet$ End-effector away from object: $  d(\mathbf{c}_i, \mathbf{y}_i) > 0 \label{eq:distswing}$, where \\ \indent \indent \(d(\cdot)\) $\in \mathbb{R}$ is the signed distance between the end-effector \\ \indent \indent and the object.

\textit{\underline{Contact mode ($k=1$)}}:\\
\indent \indent $\bullet$ Permissible contact forces: $ \psi(\mathbf{f}_i) \geq 0 \label{eq:forceManifold}$. \\
\indent \indent $\bullet$  End-effector at the contact point:   $ d(\mathbf{c}_i, g_{pt}) = 0 \label{eq:distcontact} $, \\ \indent \indent where $g_{pt} \in \mathbb{R}^{\nu}$ is the defined desired contact point on \ \indent \indent the object's surface.

Next, further design choices of this TO problem are listed.
\subsubsection*{\textbf{Permissible contact forces}}\label{subsubsec:feasibleForceManifold}

These forces should be unilateral and should lie inside the friction cone.
We choose to enforce these constraints using the vertex representation~\cite{chatzinikolaidis2018nonlinear}.

\begin{figure*}[t!]
    \centering
    \subfloat[]{\includegraphics[width=0.225\textwidth]{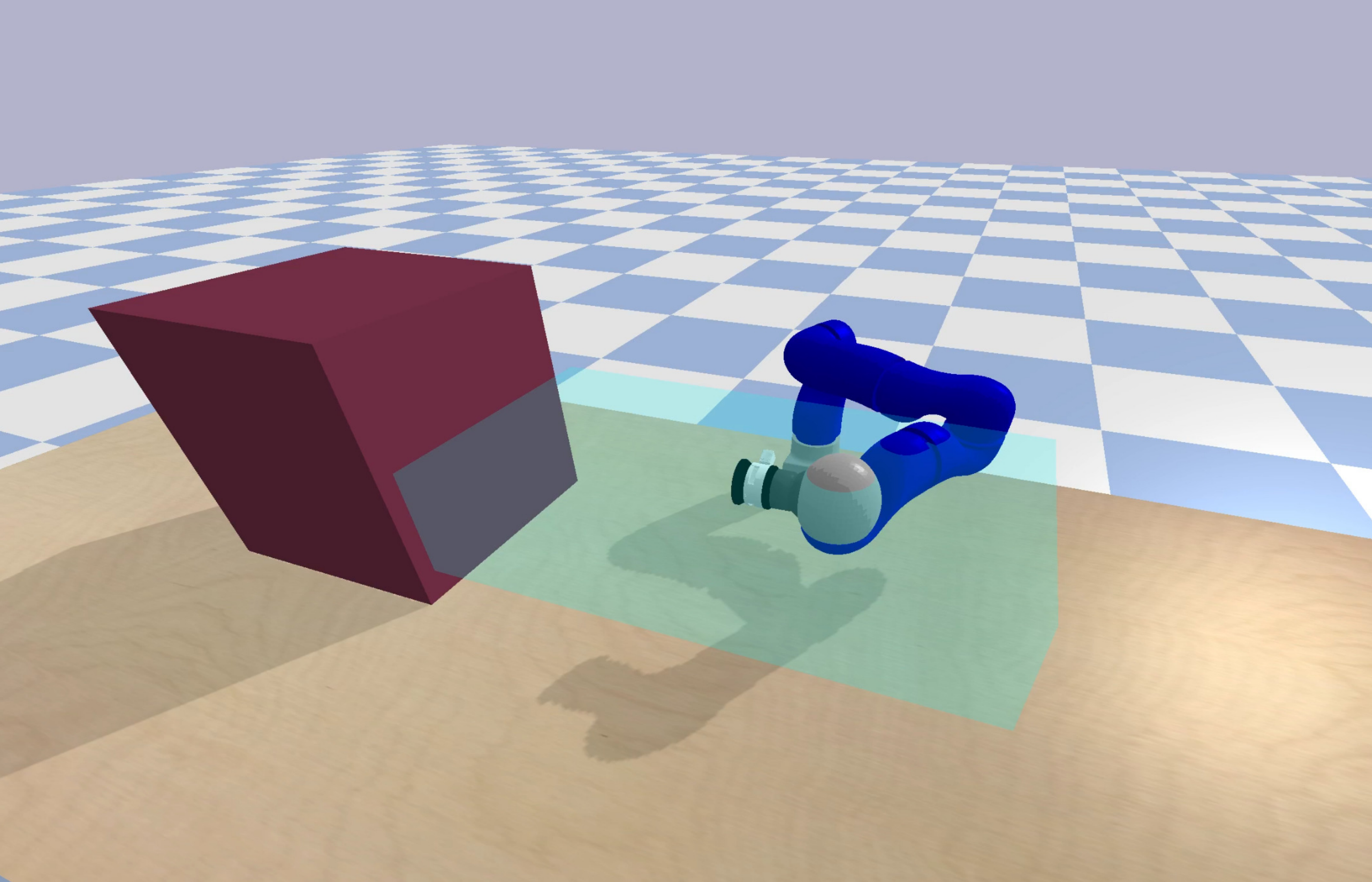}}~
    \subfloat[]{\includegraphics[width=0.225\textwidth]{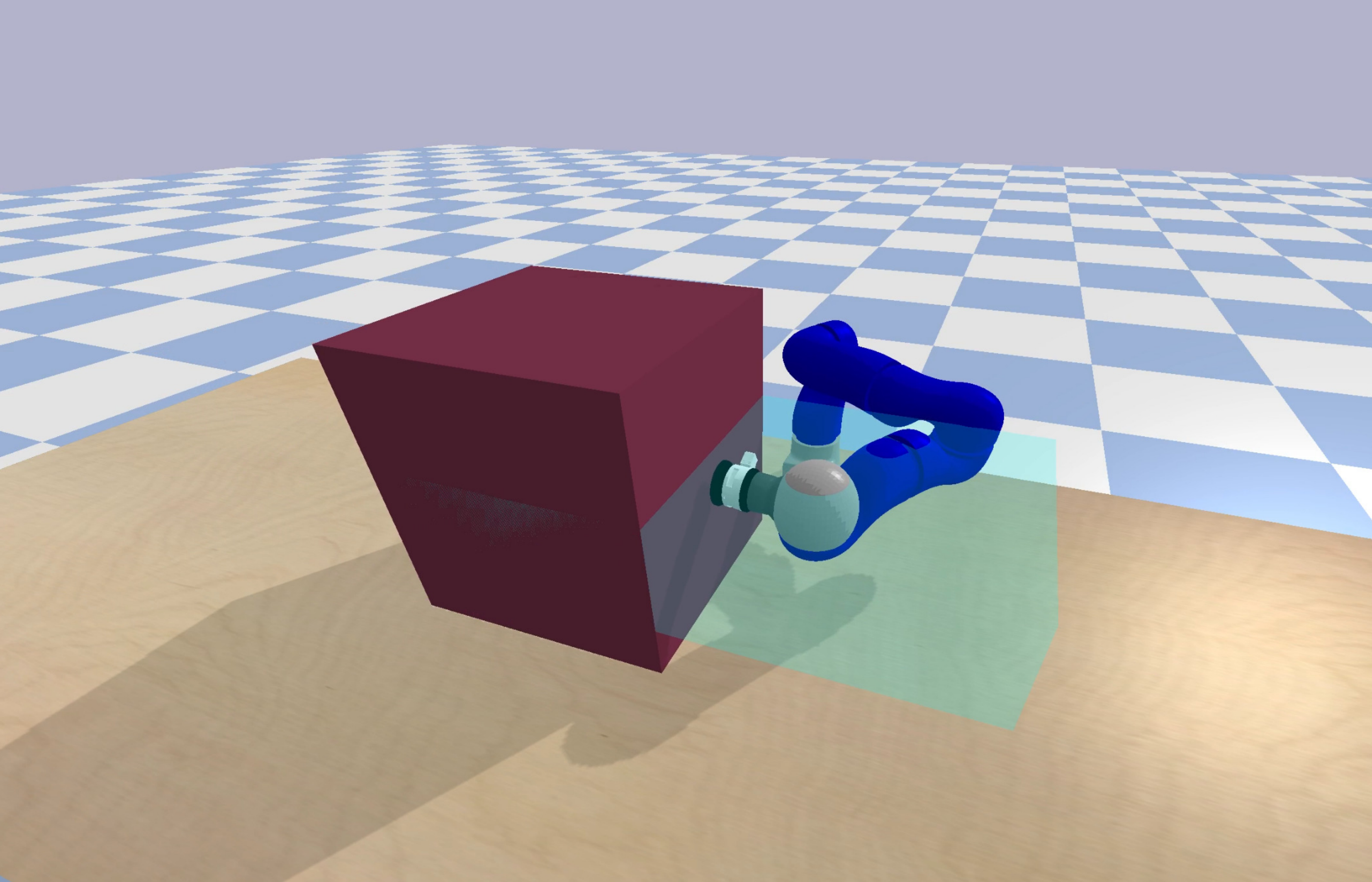}}~
    \subfloat[]{\includegraphics[width=0.225\textwidth]{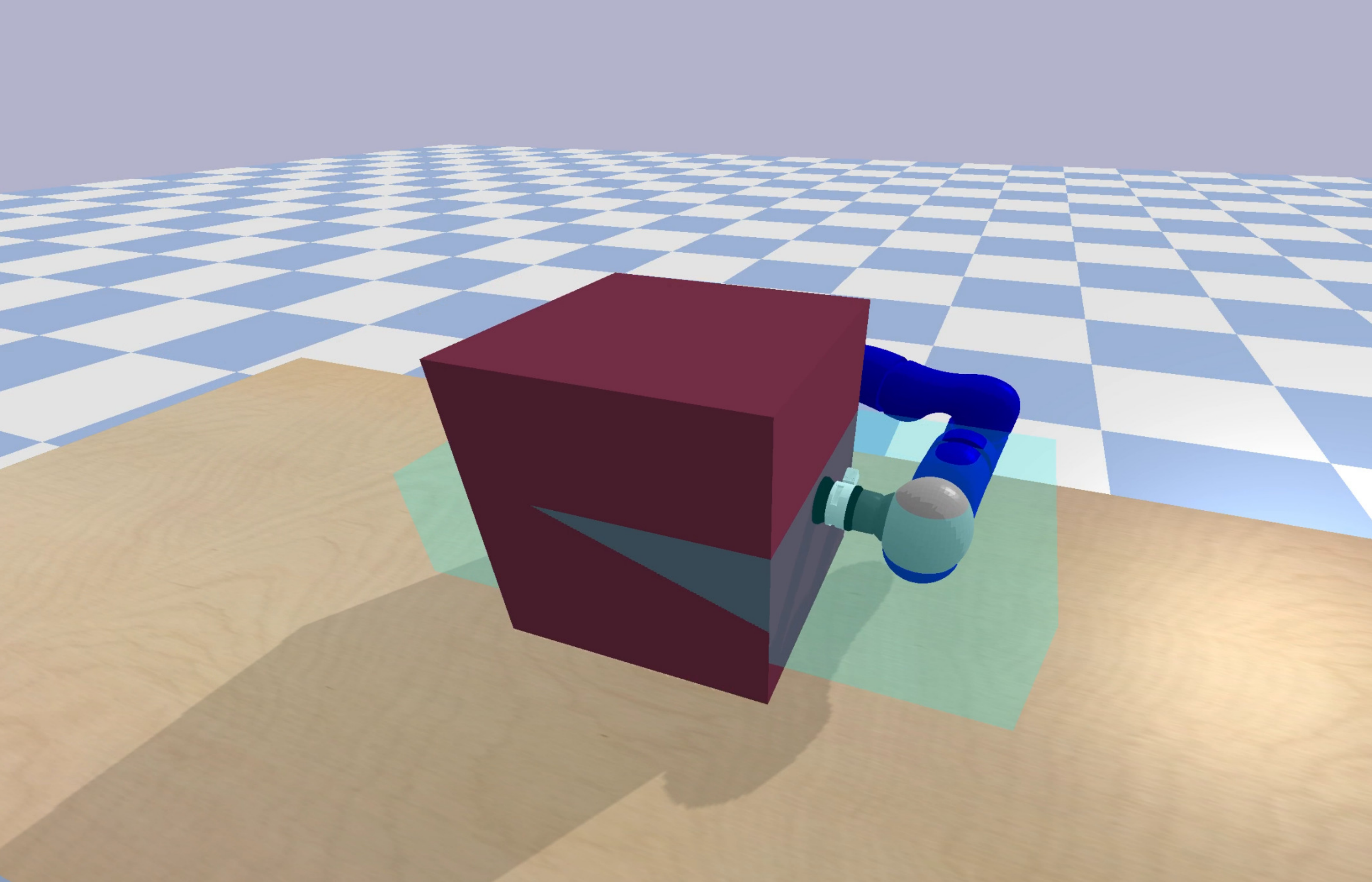}}~
    \subfloat[]{\includegraphics[width=0.225\textwidth]{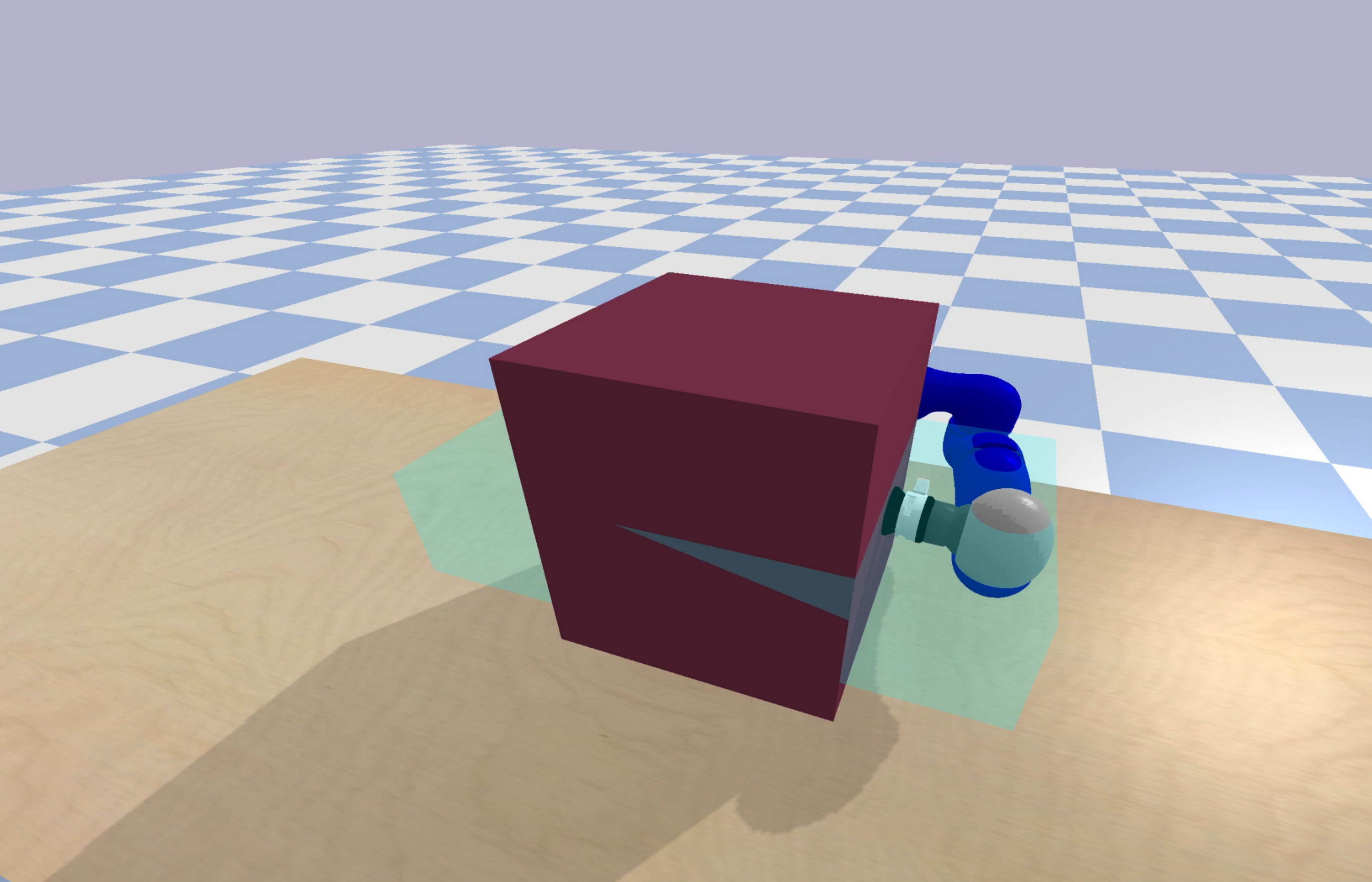}}~
    \caption{In these keyframes the robot halts a moving object travelling with 0.65 $m/s$. The cyan transparent box is an illustration of the robot's workspace.}
    \vspace{-2mm}
    \label{fig:simulation_results}
\end{figure*}

\subsubsection*{\textbf{Representations}}
The end-effector's position is represented relative to the world in Cartesian coordinates. The object's surface is represented with a closed cubic spline~\cite{stouraitis2018dyadic}.

\subsubsection*{\textbf{Input variables and hyper-parameters}}\label{subsubsec:hybridhyperparams}
The two required input variables are the sequence of contact-control modes $\bm{z}$ and the description of the manipulation task. The latter is expressed as the initial \(\begin{bmatrix} \mathbf{y}_0^* & \mathbf{\dot{y}}_0^* \end{bmatrix}\) and the goal state \(\begin{bmatrix} \mathbf{y}_N^* & \mathbf{\dot{y}}_N^* \end{bmatrix}\) of the object. 
Other type of specifications can be embedded either via the cost function $c(\cdot)$, \eg minimize duration of motion, or via the constraints $\bm{g}(\cdot)$, \eg set forbidden regions of the workspace. Further, to solve the TO problem, the number of knots $N$ needs to be specified.

%% file: sections/results.tex
In this section, we first perform a simulation ablation study on methods that could halt a moving object. We investigate whether standard compliance controllers can  reduce the contact force by tuning their compliance parameters. We also study the effects of realizing the motion plans obtained by TO methods in physical simulation. Specifically, we compare an impact-agnostic method against the impact-aware method. Next, we report computational results in which contact force profiles are optimized to satisfy both workspace limits of the robot and the task of manipulating a large-momentum object. Last, we evaluate the proposed method with real-world experiments and compare against standard compliance control. See the attached video\footnote{Link to video: https://bit.ly/338BTY1} for a physical simulation and robot experiments
of impact-aware manipulation tasks.

\paragraph*{Implementation setup} 
We use \texttt{CasADi} \cite{Andersson2018} and its automatic differentiation capabilities to realize the multi-mode TO method.
Motion planning is done in the task space and the motions are projected into the configuration space of the robot with IK. 
All simulations are conducted on a 64-bit Intel Quad-Core i9 3.60GHz computer with 64GB RAM and are realized with the Bullet physics simulation library.

\paragraph*{Simulation setup} 
The task is to halt a moving object, as shown in \cref{fig:simulation_results}. The object's mass is 20 kg and its velocity is [0.0 0.65 0.0] $m/s$. There is no sliding friction between the object and the table, while the contact is rigid.

\subsection{Standard compliance control}\label{subsec:sim_naive_method}
In the first ablation scenario, the manipulator attempts to halt the moving object only utilizing a compliant controller. The desired contact force is 0 $N$. This means that the robot will remain still till contact occurs and then, it will give in according to the selected compliance characteristics.
We explored a set of different mass-spring-damper parameters, yet independently of the parameters---as there is no pre-contact motion by the robot, the speed of the object at impact is  $\mathrm{v}^- = 0.65$~$m/s$---the resulting impact force is $\approx 680 N$ and the 
force profile is similar to ones shown in~\cref{fig:exp_066_Complianceforce}.
Such impact force can be catastrophic for the real robot and the object and may lead to a failure to effectively control the object.

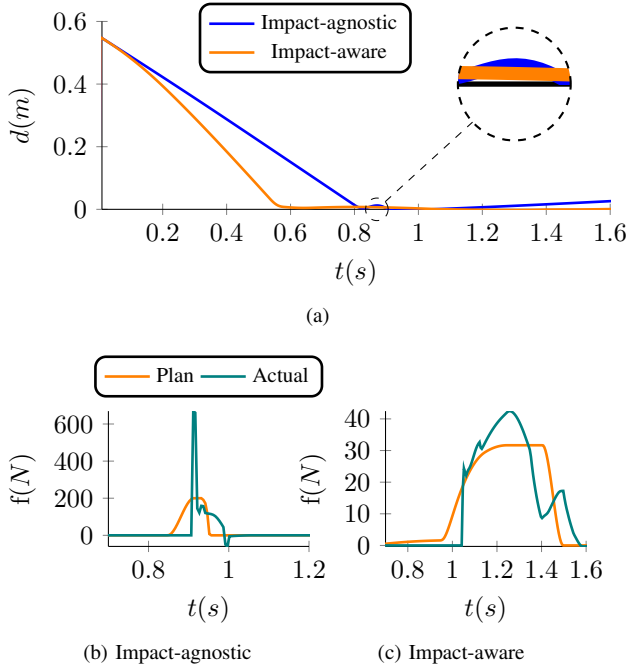
\begin{figure}[t!]
    \centering
    \subfloat[]{
        \def\svgwidth{0.95\linewidth}
        \ifdefined\figexternalized
            \tikzsetnextfilename{fig_pos_impact_aware_and_agnostic}
        \fi
        \input{figures/tikz/fig_pos_impact_aware_and_agnostic}
        \label{fig:distance_ee_vs_object_impact_cdDS} }\\   
    \subfloat[Impact-agnostic]{
        \def\svgwidth{0.40\linewidth}
        \ifdefined\figexternalized
            \tikzsetnextfilename{fig_force_impact_agnostic}
        \fi
        \input{figures/tikz/fig_force_impact_agnostic}
        \label{fig:force_impacr_sim}}
        \hspace{-12mm}
    \subfloat[Impact-aware]{
        \def\svgwidth{0.40\linewidth}
        \ifdefined\figexternalized
            \tikzsetnextfilename{fig_force_impact_aware}
        \fi
        \input{figures/tikz/fig_force_impact_aware}
        \label{fig:force_cdDS_sim}}
    \caption{Impact-agnostic versus impact-aware. Relative distance between object and the end-effector (a) for both methods.
    (b) and (c) show planed and measured in simulation contact forces for each method.}
    \label{fig:impactVScdDS}
    \vspace{-4mm}
\end{figure}

\subsection{Impact-agnostic vs Impact-aware methods}
Impact-agnostic is a term we use to refer to TO methods that plan the contact force solely on the complementarity condition~\cite{stouraitis2018dyadic, onol2019contact, sleiman2019contact}, while the impact-aware TO is realized with the multi-mode TO that utilizes the proposed contact force transmission model (see \cref{subsec:force_transmission_model}). In the second ablation study, we compare these two methods on the same halting task.
The computation time for the impact-agnostic method is on average 63~$ms$ and for the impact-aware 141~$ms$.
In \cref{fig:distance_ee_vs_object_impact_cdDS}, we show the relative distance between object and the end-effector. For the impact-agnostic method, the contact transition is abrupt and drives the object away from the manipulator ($t \approx 1.2s$). For the proposed method, the contact transition is smooth and results in stable contact.  
The reason behind the two different outcomes is the planned force profiles displayed in~\cref{fig:force_impacr_sim,,fig:force_cdDS_sim}. The impact-agnostic method plans a single impulsive force with duration $\delta t \approx 0.15s$---similar to~\cite{sleiman2019contact}---which stops the object abruptly. The planned force cannot be tracked accurately, which results in an even higher impulsive force.
In contrast, due to the enlarged duration ($\delta t \approx 0.9s$) of the planned force, our proposed method generates tractable force profiles that smoothly halt the object.
It is worth noticing the scale difference of the y axis in~\cref{fig:force_impacr_sim,,fig:force_cdDS_sim}.

\subsection{Halt-Push the object with multi-mode TO}

Here, the task is altered. In addition to the task of halting the object, the manipulator needs to push it back to a desired location. The initial object's velocity is [0.0 0.4 0.0] $m/s$. This task emphasizes the stiffness regulation capability of the proposed method. We report the optimized $\bm \alpha_l$ and contact force profile that simultaneously satisfy the task and the workspace limits of the robot. The computation times for these evaluation are between 96~$ms$ and 512~$ms$.

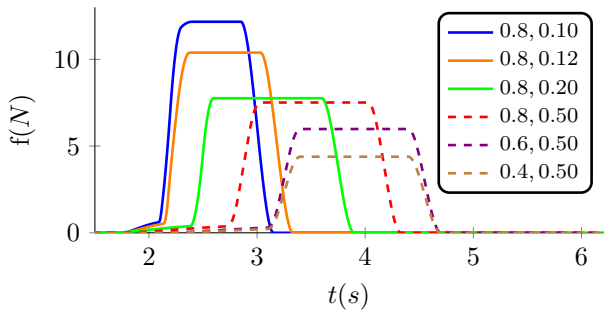
\begin{figure}[t]
\centering
\ifdefined\figexternalized
    \tikzsetnextfilename{fig_force_profile}
\fi
\input{figures/tikz/fig_force_profile}
\vspace{-12pt}
\caption{Contact force profiles with different desired positions for the object (left column) and workspace limits of the robot (right column).}
\label{fig:force_profile_diffWsPos}
\vspace{-4mm}
\end{figure}

In~\cref{fig:force_profile_diffWsPos}, we display the planned contact force profile for a number of different workspace limits (box bounds) and a number of different desired positions for the object. For the same desired position 0.8 $m$, if the workspace limit of the manipulator is increased, 
the optimized $\alpha_{-1}$ (halting stage) gradually decreases from 7.72 to 2.23, while $\alpha_1$ (pushing stage) remains at the maximum value of 20. Also, the contact duration is increased and the maximum contact force is decreased. On the other hand, by reducing the workspace limit towards zero, the planned force becomes an impulsive force--similar to impact-agnostic methods.
For a fixed workspace limit 0.5~$m$, if object's desired position is varied from 0.8~$m$ to 0.4~$m$, the maximum contact force decreases, while $\alpha_{-1}$ (halting stage) gradually decreases from 2.23 to 1.75 and $\alpha_1$ (pushing stage) remains at the upper bound value 20. In these cases, the contact duration is similar and only the force magnitude is adapted.

Furthermore, in all cases, due to the individual stiffness regulation for each contact stage based on $\alpha_l$, the planned contact force increases slowly to refrain from impacts during the halting stage ($l$ = -1), while in the pushing stage ($l$ = 1) the planned contact force increases rapidly in order to push the object to the desired position.

\subsection{Robot experiments}\label{sec:exp_results}

\begin{figure*}[t!]
    \centering
    \subfloat[]{\includegraphics[width=0.240\textwidth]{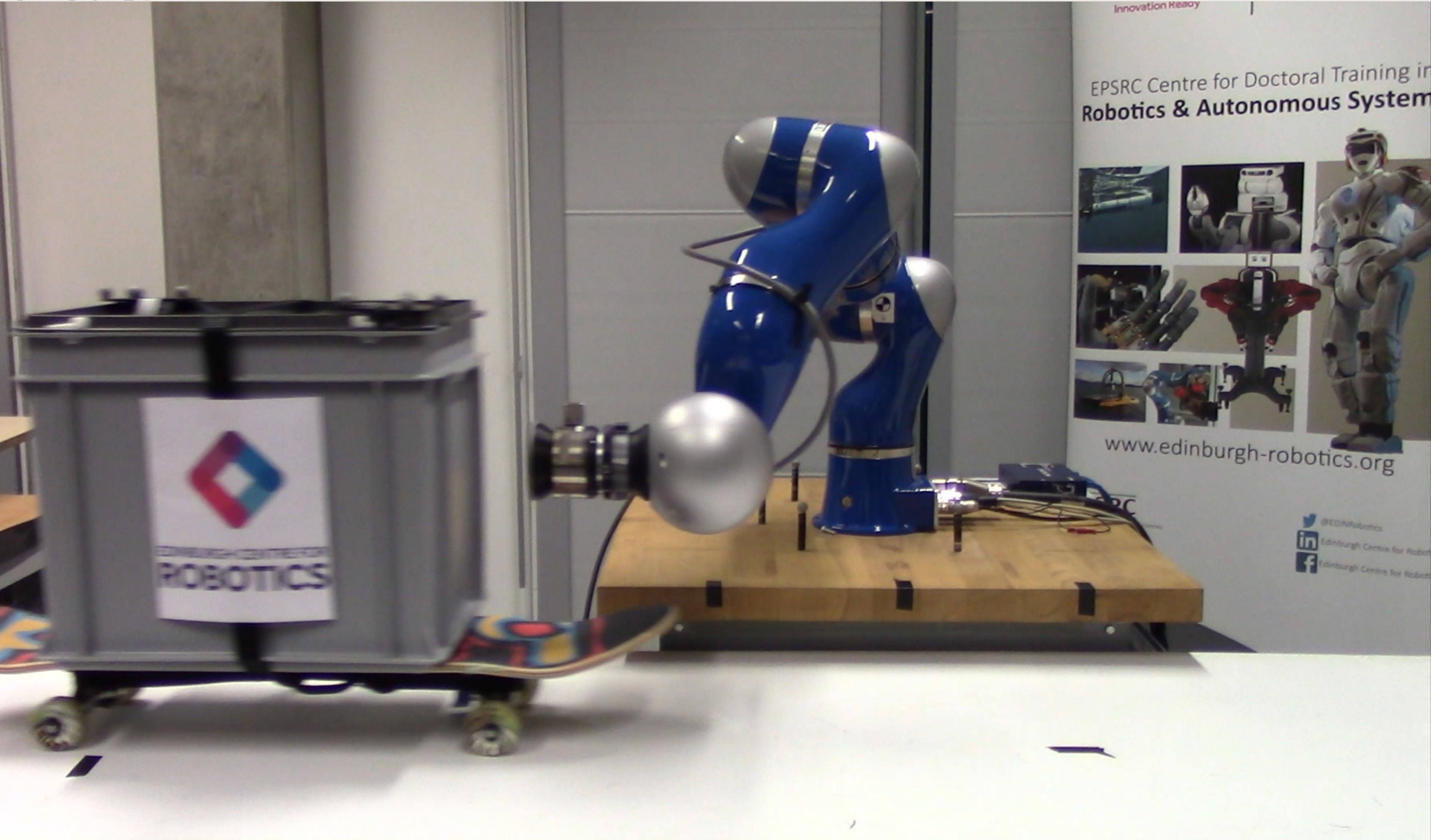}}~
    \subfloat[]{\includegraphics[width=0.240\textwidth]{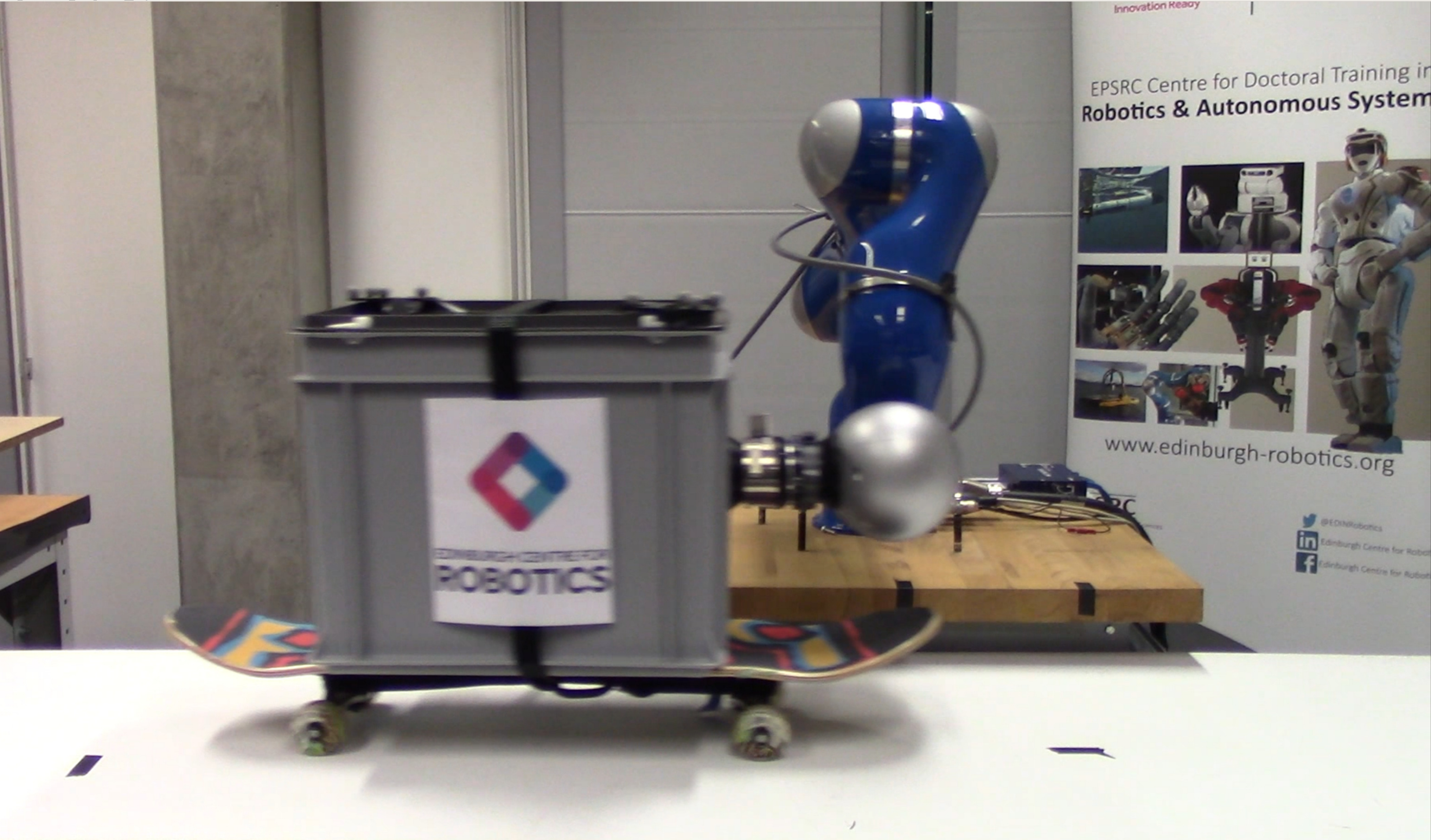}}~
    \subfloat[]{\includegraphics[width=0.240\textwidth]{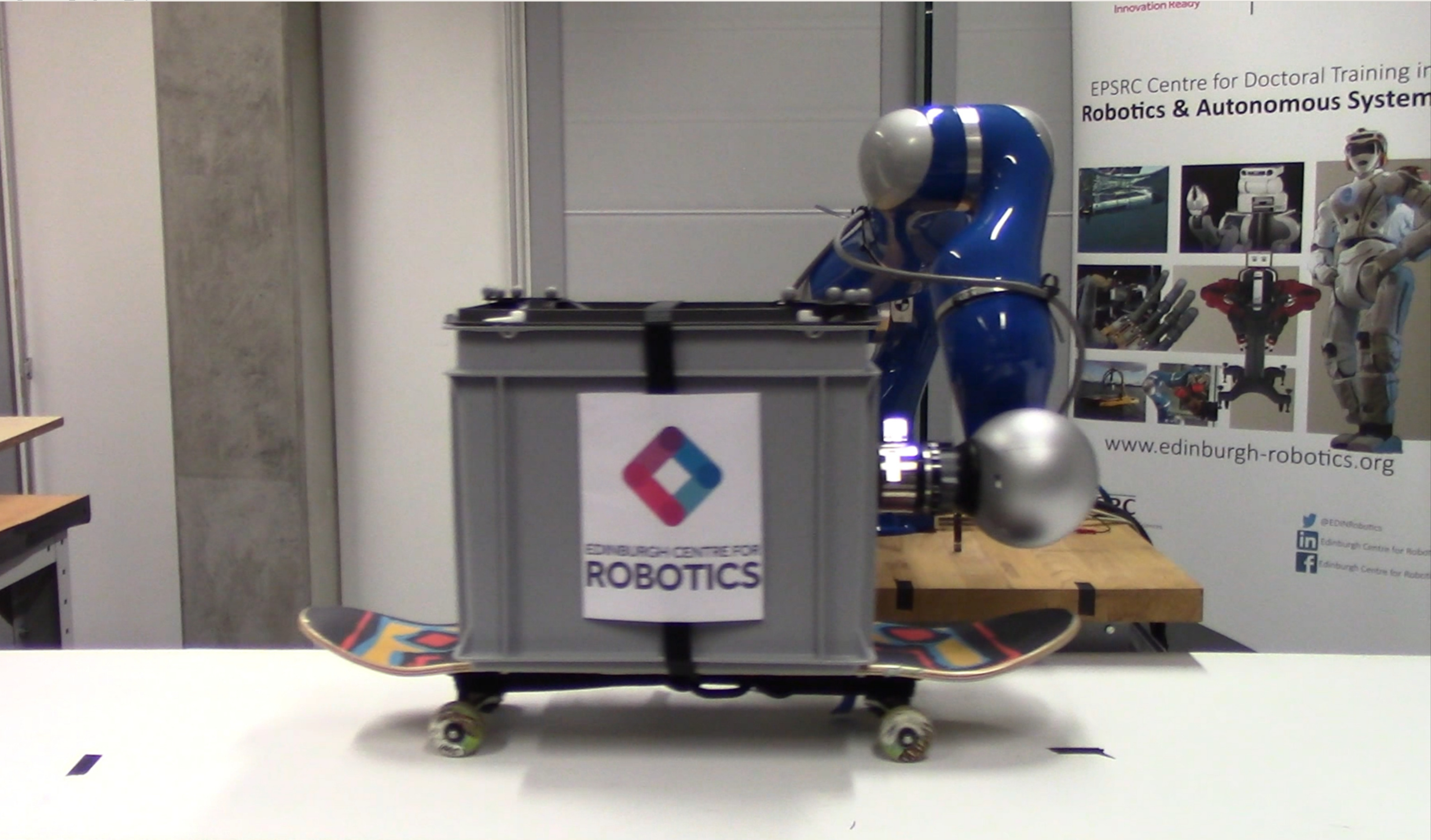}}~
    \subfloat[]{\includegraphics[width=0.240\textwidth]{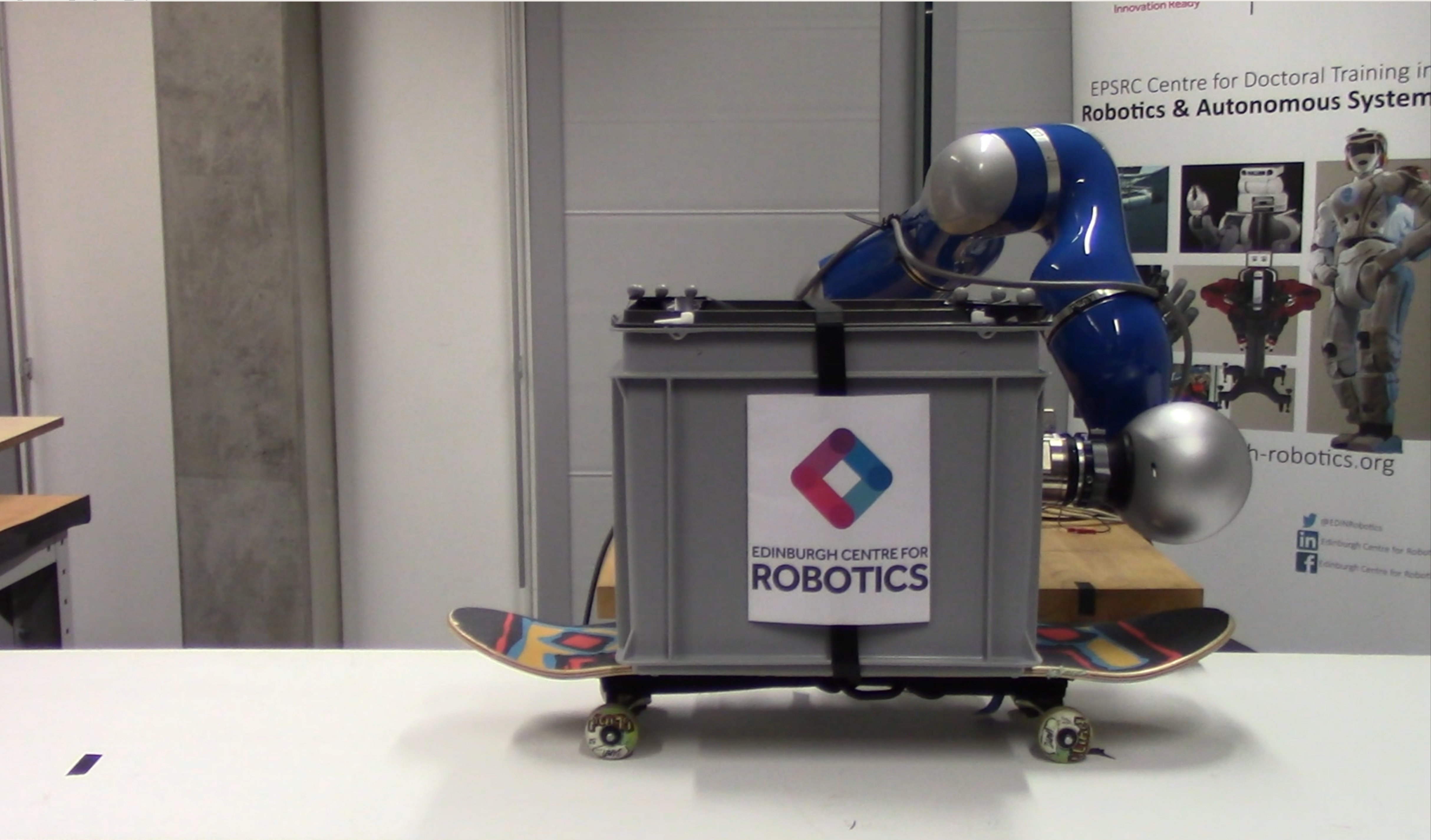}}~
    \caption{Keyframes of the experiment where the robot halts a moving object with speed of 0.66m/s.}
    \label{fig:experimental_keyframe_results}
    \vspace{-3mm}
\end{figure*}

We validate our approach in a real setting with the KUKA LWR arm and the Vicon motion capture system, where the latter is used to measure the position of the object in real time. The experimental setup is shown in~\cref{fig:teaser,,fig:experimental_keyframe_results}. 
The object is 20 $kg$ and its initial position on a slope accelerates it to an initial velocity. 
Once the object arrives in a predefined position range---based on measurements of its position---we estimate its velocity and acceleration on-the-fly. These are estimated 
by fitting a rolling friction model with the \textit{coefficient of determination} 
being in the range $0.5 \leq r^2 \leq 0.8$, which implies 
that there is uncertainty in the motion estimation.
These estimated values are then passed on to the impact-aware method, which predicts (by integration) the future motion of the object, and computes an optimal motion plan in less than 150 $ms$ to halt the  object within the workspace limits. The position and stiffness profiles of the motion plan are streamed to the robot, such that, the joint position with Cartesian stiffness control mode of the KUKA LWR arm tracks the optimal motion.
The impact-aware position trajectory is optimized in the task space and is realized on the robot in an open-loop fashion after being mapped to the configuration space using IK, along with the task space stiffness trajectory. The manipulator is controlled at 200 $Hz$, while the Vicon tracking runs at 100 $Hz$. The material of end-effector is stiff to have rigid contact.

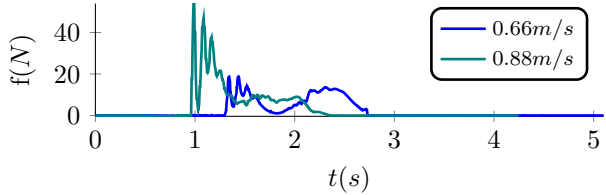
\begin{figure}[t]
\centering
\ifdefined\figexternalized
    \tikzsetnextfilename{fig_force_66and88}
\fi
\input{figures/tikz/fig_force_66and88}
\vspace{-12pt}
\caption{Experimental result of contact force during halting motion.}
\label{fig:exp_force}
\vspace{-2mm}
\end{figure}

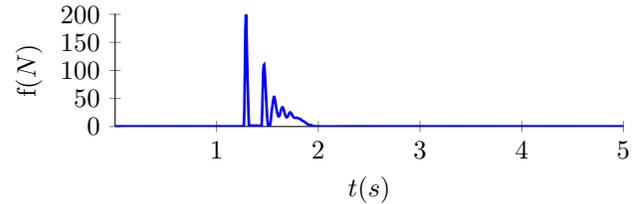
\begin{figure}[t]
\centering
\ifdefined\figexternalized
    \tikzsetnextfilename{fig_compliant_impact_66}
\fi
\input{figures/tikz/fig_compliant_impact_66}
\vspace{-12pt}
\caption{Experimental result of impact between the object and the end-effector during halting an object with speed of 0.66 $m/s$.}
\label{fig:exp_066_Complianceforce}
\vspace{-5mm}
\end{figure}

In~\cref{fig:exp_force}, we report the measured contact force 
obtained by an ATI F/T sensor at 100~$Hz$. The object travelled at the speed of approx. 0.66~$m/s$ and the  proposed method halts its motion with a maximum force less than 20~$N$ in a full workspace scenario, and less than 30~$N$ with reduced workspace (see video).
As a baseline---similarly to~\cref{subsec:sim_naive_method}---we report the measured force with a very soft configuration ($K = 10$ and $\lambda = 1$) of the LWR arm's compliance controller. In this case, the maximum impact force is 199.47~$N$ (see \cref{fig:exp_066_Complianceforce}), which is 10 times larger than the one shown in~\cref{fig:exp_force}.

\begin{figure}[t]
\centering
\ifdefined\figexternalized
    \tikzsetnextfilename{fig_pos_88}
\fi
\input{figures/tikz/fig_pos_88}
\vspace{-12pt}
\caption{Experimental result of the position of the object and the end-effector during halting an object with speed of 0.88 $m/s$.}
\label{fig:exp_088_position}
\vspace{-4mm}
\end{figure}
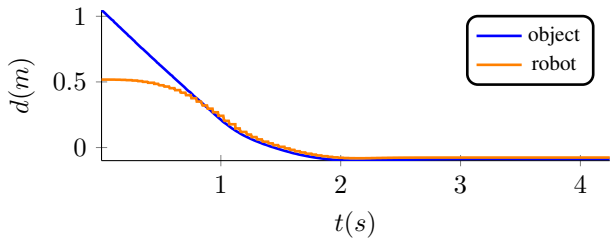

Furthermore, to emphasize the capabilities of the method, we consider the same object with a speed of 0.88~$m/s$. The positions of the object and the end-effector 
are shown in \cref{fig:exp_088_position} to display their alignment, 
while the contact force is spread-out as shown in~\cref{fig:exp_force}.
For these initial conditions the contact force remains smaller than 55~$N$, while the baseline is omitted in order to avoid 
stressing the hardware.

%% file: figures/tikz/fig_pos_impact_aware_and_agnostic.tex
\begin{tikzpicture}[spy using outlines=
	{circle, magnification=5, connect spies},
	]
    \begin{axis}[
        xlabel={$t(s)$},
        ylabel={$d(m)$},
        height=2.5cm,
        xmin=0.01,xmax=1.6,
        ymin=0, ymax=0.6,
        legend style={at={(0.4,1.1)},anchor=north},
        ]
        \addplot[
            no markers,
            blue,
        ] table [
            col sep=comma,
            x={time},
            y={pos_error}
        ] {figures/data/force_impact_agnostic.csv};
        \addlegendentry{Impact-agnostic}
        \addplot[
            no markers,
            orange,
        ] table [
            col sep=comma,
            x={time},
            y={pos_error}
        ] {figures/data/force_impact_aware.csv};
        \addlegendentry{Impact-aware}
        \coordinate (spypoint) at (axis cs:0.87,0.0);
        \coordinate (magnifyglass) at (axis cs:1.3,0.4);
    \end{axis}
    \spy [black, dashed, size=1.5cm] on (spypoint) in node[fill=white] at (magnifyglass);
\end{tikzpicture}

%% file: figures/tikz/fig_force_impact_agnostic.tex
\begin{tikzpicture}
    \begin{axis}[
        xlabel={$t(s)$},
        ylabel={$\mathrm{f}(N)$},
        height=1.8cm,
        xmin=0.7,xmax=1.2,
        width=0.15\textwidth,
        legend columns=2,
        every axis legend/.append style={
            at={(0.5,1.1)},
            anchor=south
        },
        ]
        \addplot[
            no markers,
            orange,
        ] table [
            col sep=comma,
            x={time},
            y={force_plan}
        ] {figures/data/force_impact_agnostic.csv};
        \addlegendentry{Plan}
        \addplot[
            no markers,
            teal,
        ] table [
            col sep=comma,
            x={time},
            y={force_measured}
        ] {figures/data/force_impact_agnostic.csv};
        \addlegendentry{Actual}
    \end{axis}
\end{tikzpicture}

%% file: figures/tikz/fig_force_impact_aware.tex
\begin{tikzpicture}
    \begin{axis}[
        xlabel={$t(s)$},
        ylabel={$\mathrm{f}(N)$},
        height=1.8cm,
        xmin=0.7,xmax=1.6,
        width=0.15\textwidth,
        legend columns=2,
        every axis legend/.append style={
            at={(0.5,1.1)},
            anchor=south
        },
        ]
        \addplot[
            no markers,
            orange,
        ] table [
            col sep=comma,
            x={time},
            y={force_plan}
        ] {figures/data/force_impact_aware.csv};
        \addplot[
            no markers,
            teal,
        ] table [
            col sep=comma,
            x={time},
            y={force_measured}
        ] {figures/data/force_impact_aware.csv};
    \end{axis}
\end{tikzpicture}

%% file: figures/tikz/fig_force_profile.tex
\begin{tikzpicture}
    \begin{axis}[
        xlabel={$t(s)$},
        ylabel={$\mathrm{f}(N)$},
        height=3.0cm,
        xmin=1.5,xmax=6.2,
        ymin=0,ymax=13,
        ]
        \addplot[
            no markers,
            blue,
        ] table [
            col sep=comma,
            x={time},
            y = {force},
        ] {figures/data/profile_08_005.csv};
        \addlegendentry{$0.8, 0.10$}
        \addplot[
            no markers,
            orange,
        ] table [
            col sep=comma,
            x={time},
            y = {force},
        ] {figures/data/profile_08_006.csv};
        \addlegendentry{$0.8, 0.12$}
        \addplot[
            no markers,
            green,
        ] table [
            col sep=comma,
            x={time},
            y = {force},
        ] {figures/data/profile_08_01.csv};
        \addlegendentry{$0.8, 0.20$}
        \addplot[
            no markers,
            red,
            dashed
        ] table [
            col sep=comma,
            x={time},
            y = {force},
        ] {figures/data/profile_08_025.csv};
        \addlegendentry{$0.8, 0.50$}
        \addplot[
            no markers,
            dashed,
            violet,
        ] table [
            col sep=comma,
            x={time},
            y = {force},
        ] {figures/data/profile_06_025.csv};
        \addlegendentry{$0.6, 0.50$}
        \addplot[
            no markers,
            dashed,
            brown,
        ] table [
            col sep=comma,
            x={time},
            y = {force},
        ] {figures/data/profile_04_025.csv};
        \addlegendentry{$0.4, 0.50$}
    \end{axis}
\end{tikzpicture}

%% file: figures/tikz/fig_force_66and88.tex
\begin{tikzpicture}
    \begin{axis}[
        xlabel={$t(s)$},
        ylabel={$\mathrm{f}(N)$},
        height=1.5cm,
        legend pos=north east,
        ]
        \addplot[
            no markers,
            blue,
        ] table [
            col sep=comma,
            x={time},
            y expr={-\thisrow{force}}
        ] {figures/data/force_bag_impact_opt_20kg_066ms_i5_good.dat};
        \addlegendentry{$0.66 m/s$}
        \addplot[
            no markers,
            teal,
        ] table [
            col sep=comma,
            x={time},
            y expr={-\thisrow{force}}
        ] {figures/data/force_bag_impact_opt_20kg_088ms_i10.dat};
        \addlegendentry{$0.88 m/s$}
    \end{axis}
\end{tikzpicture}

%% file: figures/tikz/fig_pos_88.tex
\begin{tikzpicture}
    \begin{axis}[
        xlabel={$t(s)$},
        ylabel={$d(m)$},
        height=2cm,
        legend pos=north east,
        ]
        \addplot[
            no markers,
            blue,
        ] table [
            col sep=comma,
            x={time},
            y expr={-\thisrow{pos}}
        ] {figures/data/pos_obj_bag_impact_opt_20kg_088ms_i10.dat};
        \addlegendentry{object}
        \addplot[
            no markers,
            orange,
        ] table [
            col sep=comma,
            x={time},
            y expr={-\thisrow{pos}}
        ] {figures/data/pos_robot_bag_impact_opt_20kg_088ms_i10.dat};
        \addlegendentry{robot}
    \end{axis}
\end{tikzpicture}

%% file: figures/tikz/fig_compliant_impact_66.tex
\begin{tikzpicture}
    \begin{axis}[
        xlabel={$t(s)$},
        ylabel={$\mathrm{f}(N)$},
        height=1.5cm,
        ymax=201,
        ]
        \addplot[
            no markers,
            blue,
        ] table [
            col sep=comma,
            x={time},
            y expr={-\thisrow{force}}
        ] {figures/data/compliant_impact_force.dat};
    \end{axis}
\end{tikzpicture}

%% file: sections/conclusion.tex
This paper presents a novel impact-aware multi-mode Trajectory Optimization method that encodes both hybrid dynamics and hybrid control in a single formulation.
This enables planning smooth transitions from free-motion to contact at speed. 
The method uses the proposed contact force transmission model to compute the optimal halting motion and stiffness profile that respect hardware's limitations such as force and workspace limits. 
We evaluated the method both in simulation and real-world experiments for a task of halting a large mass and fast moving object with a compliant robot.
Both results demonstrate that the proposed method enables much lower contact transition forces than standard compliance controllers and impact-agnostic TO methods.

Currently the robot computes the impact-aware trajectories online, but executes them in open-loop, requiring fairly accurate estimation of motion parameters such as velocity and rolling friction. Future aspirations include extending the framework to a bimanual setup, towards manipulating 
flying and tumbling objects. To realize this aspiration, we will explore extending the proposed framework to a fully closed loop MPC implementation.